\documentclass[pdflatex,sn-mathphys-ay]{sn-jnl} 
\usepackage{graphicx}%
\usepackage{multirow}%
\usepackage{amsmath,amssymb,amsfonts}%
\usepackage{amsthm}%
\usepackage{mathrsfs}%
\usepackage[title]{appendix}%
\usepackage{xcolor}%
\usepackage{textcomp}%
\usepackage{manyfoot}%
\usepackage{booktabs}%
\usepackage{algorithm}%
\usepackage{algorithmicx}%
\usepackage{algpseudocode}%
\usepackage{listings}%

\theoremstyle{thmstyleone}%

\theoremstyle{thmstyletwo}%

\theoremstyle{thmstylethree}%

\begin{document}

\title[]{Spatiotemporal Graph Transformer for 3D Neighborhood Interaction and Quality Prediction in Metal Additive Manufacturing}

\author[1]{\fnm{Joyce Karen} \sur{Pelaez}}
\author*[1]{\fnm{Siqi} \sur{Zhang}}\email{sizhang@siue.edu}
\author[1]{\fnm{Hoo Sang} \sur{Ko}}

\affil[1]{\orgdiv{Department of Industrial Engineering}, \orgname{Southern Illinois University Edwardsville}, \orgaddress{\street{Box 1805}, \city{Edwardsville}, \postcode{62026}, \state{IL}, \country{USA}}}

\abstract{Metal additive manufacturing enables the fabrication of complex parts, but achieving consistent build quality remains challenging due to interactions induced by repeated layer-wise melting, solidification, and reheating across the 3D build. Advanced sensing provide a great opportunity to collect rich observations of the actual manufacturing process for real-time quality monitoring and control. Yet, existing methods often have limited ability to represent multi-layer interactions and quantify their contributions to quality. In this paper, we develop a novel spatiotemporal graph transformer for modeling 3D neighborhood interactions and learn their effects on build quality in metal additive manufacturing. Specifically, we first introduce a weighted network representation of the manufacturing process, where fusing locations are modeled as nodes, and their spatial- and process-dependent relationships are encoded as edge weights. This representation also enables the integration of multimodal data (e.g., geometric design, process settings, and in-situ sensing data) into a unified structure for downstream learning tasks. Building on this network, we further design a dual-attention graph transformer that captures both within-node feature dependencies and cross-node neighborhood interactions for quality representation learning. Experimental results show that the proposed framework significantly outperforms image-based, sequence-based, and graph-based models in characterizing process-quality relationships. More importantly, the incorporation of cross-layer interactions is critical for improving quality prediction performance. This framework is broadly applicable to other tasks involving network modeling and graph-based representation learning.}


\maketitle

\section{Introduction} \label{sec1:intro}
Metal additive manufacturing (MAM) has emerged as an important manufacturing paradigm for producing complex, customized and high-performance components beyond the capabilities of conventional manufacturing processes. However, ensuring consistent as-built quality and performance remains a persistent challenge and a key barrier to its widespread adoption. During the layerwise fabrication process, repeated melting, solidification, and reheating introduce complex thermal interactions throughout the three-dimensional build. These interactions can accumulate across layers and lead to quality variations or defect formation \citep{grasso2017process}. Such layer-to-layer dependencies make it difficult to characterize process-quality relationships, especially under heterogeneous geometric and process settings. Establishing these relationships is central to process-structure-property understanding and provides the basis for more reliable quality monitoring and control. 

Recent advances in sensing modalities and data acquisition have created new opportunities to establish process-quality relationships in MAM. In the state of the art, in-situ monitoring systems can capture rich process signatures during fabrication, including optical imaging of layerwise finishes \citep{abdelrahman2017flaw}, melt-pool images related to microstructure evolution \citep{khanzadeh2019situ}, electron emission associated with laser-material interaction dynamics \citep{depond2020laser}, and acoustic responses associated with mechanical transients \citep{yang2020six}. Broadly, these data streams characterize build quality from two complementary perspectives: layerwise observations that describe process behavior across an entire build layer and pointwise observations collected at individual fusing locations. Layerwise observations are valuable for identifying surface anomalies and spatially distributed defects over each fabricated layer. In contrast, pointwise observations are important for quality monitoring because they provide location-specific information about local thermal and material response during fabrication, which is closely related to microstructure evolution and defect formation. 

Machine learning and deep learning models have shown strong potential for extracting useful information from high-dimensional sensing data and supporting quality monitoring in MAM. Nonetheless, many data-driven models primarily learn statistical associations from observed data, without explicitly representing multi-layer interactions underlying quality formation. This limitation is particularly important for pointwise quality analysis, where the response at a given fusing location depends not only on the local process signal, but also on its surrounding process history. During fabrication, neighboring scan tracks and  nearby fusing locations in previously deposited layers can jointly influence the local material conditions and quality outcomes. Hence, pointwise quality prediction requires modeling approaches that explicitly represent multi-layer process interactions and neighborhood history. 

However, modeling multi-layer process interactions is not straightforward because these interactions are often not uniform across the three-dimensional build. Different regions of a part can experience distinct local geometries, scan strategies, and process settings, especially when complex geometric features are involved. As a result, the influence of neighboring scan tracks and previously deposited layers on local quality can vary depending on the spatial and process context of each fusing location. This calls for a modeling framework that can represent heterogeneous relationships among fusing points within and across layers and learn quality-relevant representations from their neighborhood contexts.

This work addresses these challenges by developing a spatiotemporal graph transformer for 3D neighborhood modeling and quality prediction in metal additive manufacturing. First, a weighted network representation is introduced to connect fusing locations throughout the 3D build and characterize their heterogeneous relationships based on spatial coordinates and process settings. Second, a spatiotemporal graph transformer is proposed to learn quality-relevant representations from the constructed weighted network. The core component of the transformer is a dual-attention mechanism that enables both within-node feature learning and cross-node interaction modeling over spatiotemporal neighborhoods. The main contributions of this work are summarized as follows, 
\begin{enumerate}
    \item \textit{Weighted Network Representation}: This work introduces a weighted network representation to model fusing locations throughout the 3D build. In this representation, fusing locations are treated as nodes and their heterogeneous relationships induced by part geometry and process settings are encoded through edge weights. This representation also enables in-situ sensing data to be directly embedded within the network structure, thereby supporting downstream process-quality learning. 
    \item \textit{Novel Spatiotemporal Transformer Model}: This work develops a spatiotemporal graph transformer (STGT) for learning quality-relevant representations from the constructed weighted network. Specifically, STGT relies on a two-level attention structure for process-quality learning: 1) within-node attention to extract local features and 2) cross-node attention to capture neighborhood interactions among connected fusing locations. By integrating these two levels, the proposed framework learns both local process signatures and spatiotemporal interaction patterns that are important for quality prediction. 
    \item \textit{In-situ Quality Prediction}: By integrating process data available during fabrication with physics-informed neighborhood relationships, the proposed framework predicts ex-situ quality responses and supports spatially resolved quality monitoring in MAM. 
\end{enumerate}

The remainder of this paper is organized as follows. Section \ref{sec:background} reviews relevant studies, including image-guided quality monitoring, graph-based modeling, and graph transformer. Section \ref{sec:methodlogy} discusses the proposed methodology. Section \ref{sec:exp design} provides the experimental design and Section \ref{sec:results} discusses the results. Finally, conclusions are drawn in Section \ref{sec:conclusions}. 

\section{Research Background}\label{sec:background}
\subsection{Data-driven Quality Monitoring}
Advanced sensing significantly enhances process observability and allows for more effective characterization of process stability and part quality. However, realizing this potential requires analytical methods that can transform high-dimensional sensing data into useful information. Traditional statistical quality control (SPC) methods typically focus on feature engineering and statistical analysis. For imaging data, this often involves extracting low-dimensional features and utilizing them as inputs for regression, classification or statistical hypothesis tests. For example, \cite{yao2018multifractal} proposed to extract fractal patterns from layerwise imaging data and perform Hotelling $T^2$ tests for the detection of AM defects. \cite{yang2023multi} developed a multi-resolution analysis based on 2D continuous wavelet transform to delineate defect characteristics from AM imaging profiles and enable in-situ quality monitoring via predictive modeling. In addition, \cite{zhang2025multiscale} presented a basis function based model to derive 3D melt-pool characteristics for anomaly detection in additive manufacturing. These studies often provide good interpretability because of the nature of feature engineering, but their effectiveness is to some extent constrained by underlying assumptions and reliance on domain expertise. 

Recently, deep learning models have been increasingly explored by additive manufacturing researchers. One key characteristic is their ability to automate feature learning and enable end-to-end modeling directly from raw inputs to outputs. For example, \cite{yang2019investigation} trained a convolutional neural network to classify melt-pool images into different size categories. \cite{zhang2024engineering} incorporated metric learning into convolutional autoencoders for domain-informed feature learning. Beyond convolutional neural networks, \cite{wang2025transformer} investigated the feasibility of a vision transformer for defect classification in fused filament fabrication. Although both statistical and deep learning methods have demonstrated the value of advanced sensing data for quality monitoring, these studies tend to overlook the relational structures (e.g., scan path dependencies, and layerwise interactions) that are inherent in additive manufacturing processes. 

To address the limitation, several studies have turned to spatiotemporal models for better characterizing complex process dynamics. For example, \cite{larsen2022deep} investigated a hybrid framework of a variational autoencoder and recurrent neural networks for characterizing spatiotemporal behavior in melt-pool imaging data. \cite{ogoke2024deep} developed a hybrid CNN-Transformer architecture to predict melt-pool depth contour from a temporal sequence of thermal images. In addition to point-level images, \cite{li2024time} designed a spatiotemporal vision transformer to assist fault diagnosis from layerwise imaging data in fused deposition modeling. In these studies, imaging sequences are organized according to temporal order. This works well for layerwise imaging data, where spatial and temporal relationships are consistently coupled. However, point-level images often exhibit more heterogeneous spatial and temporal dependencies. For example, two fusing locations from different layers are spatially related, even though they are separated by a large temporal gap. Such heterogeneous dependencies call for a modeling framework that is sufficiently flexible to represent nonuniform process interactions and learn process-quality relationships. 

\subsection{Graph-based Representation and Learning}

Graphs provide a natural representation of interconnected entities, in which the entities are modeled as nodes and their relationships are encoded as edges connecting pairs of nodes. In many engineering contexts, such graph representations are referred to as complex networks. Complex networks have been widely used to study complex systems across various domains, including smart factories \citep{lee2023digital}, epidemic spread \citep{zhang2021spatial} and curriculum design \citep{yang2024comparative}. Interestingly, some researchers have also modeled images as complex networks. One representative approach is to represent images at the pixel level, where individual pixels are treated as nodes \citep{kan2017dynamic, chen2018recurrence}. 

Once the graph representation is established, various analyses can be conducted to extract process-relevant information. One strand of research centers on classical complex network theory. For example, \cite{kan2017dynamic} derived community-based network statistics for monitoring and controlling imaging streams from ultraprecision machining and biomanufacturing processes. \cite{chen2018recurrence} proposed to construct networks via recurrence theorem and showed that recurrence network measures, such as average degree, path length, and density are important for evaluating surface finish quality in ultraprecision machining. 
Another strand of research focuses on graph neural networks (GNN) for automatic representation learning from graph-structured data. For example, \cite{mozaffar2021geometry} investigated GNNs for thermal modeling of additive manufacturing processes. \cite{hussong2025selection} proposed a graph-attention-network-based framework for semantic segmentation of 3D CAD models. \cite{alenezi2025graph} developed a graph-based variation propagation network for modeling multi-stage manufacturing systems. 

Although existing studies have demonstrated promising performance across different manufacturing tasks, they primarily focus on learning from network structures or topology-derived information. In other words, these methods cannot adequately assimilate in-situ sensing data into graph-based learning. This highlights the need for a graph-based learning framework that can effectively learn quality-relevant representations from in-situ sensing data over the network structure. 

\subsection{Graph Transformer}
The Transformer is a neural network architecture originally developed for sequence modeling in natural language processing \citep{vaswani2017attention}. Its core idea is to represent each input element (e.g, a word in a sentence) as a token and then employ self-attention to learn how strongly each token is related to the others. In this way, the model can capture contextual dependencies and learn informative feature representations. Graph transformers extend this idea to graph-structured data by treating nodes as tokens and modifying self-attention so that node interactions are informed by the underlying graph structure. \cite{dwivedi2020generalization} presented an early general formulation of graph transformers by incorporating graph connectivity into attention and extending the architecture to handle both node and edge features. Later, \cite{chen2022structure} further emphasized the importance of structural information and proposed a structure-aware self-attention mechanism that explicitly incorporates subgraph representations. Together, these studies highlight the potential of graph transformers as a flexible and expressive framework for learning from graph-structured data. 

Recent work has begun to explore the attention-based architectures in additive manufacturing. For instance, \cite{wang2025cross} showed that transformer-based models can achieve good performance in predicting the energy demand of printing intricate structures. \cite{uhrich2025mpgt} proposed a multimodal physics-constrained graph transformer for hybrid digital twins in additive manufacturing. Despite these advances, process-aware graph transformer frameworks for manufacturing quality monitoring remain underexplored. Little has been done to address how graph transformers can be designed to learn from spatiotemporal neighborhood interactions for effective quality prediction of MAM processes. 

\section{Research Methodology}\label{sec:methodlogy}
\begin{figure*}[ht]
    \centering
    \includegraphics[width=\linewidth]{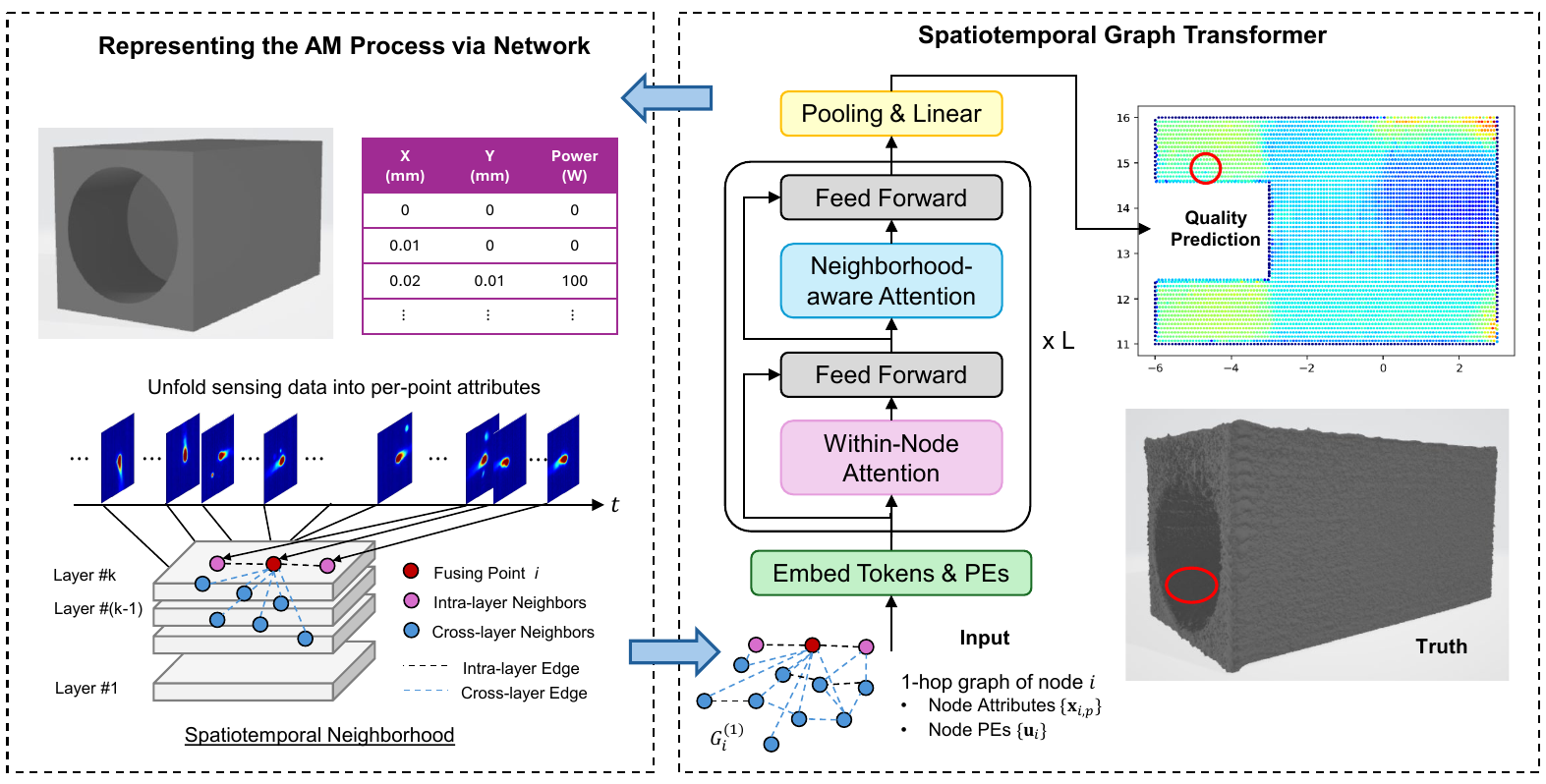}
    \caption{The proposed framework to represent and learn the AM process via spatiotemporal networks.}
    \label{fig:flowdiagram}
\end{figure*}

Figure \ref{fig:flowdiagram} presents the proposed framework for 3D neighborhood modeling and quality prediction in MAM. The key idea is to represent 3D process interactions via a weighted network and then to leverage this representation for process-quality learning. The overall framework consists of two main components: 1) network modeling of the MAM process and 2) spatiotemporal graph transformer for learning process-quality relationships over the constructed MAM process network. Unlike image-based or sequence-based models, the network representation connects neighboring fusing points in 3D space while directly incorporating their geometric and process characteristics into the graph structure. This is particularly important in MAM, where complex part geometries and layer-wise fabrication create heterogeneous spatiotemporal process interactions. The second component introduces a dual-attention learning mechanism to capture both within-node feature dependencies and cross-node neighborhood dependencies over the spatiotemporal network. Together, these two components provide a foundation for systematically learning how neighborhood dynamics across build layers influence final part quality in layer-wise manufacturing processes.

\subsection{Network Modeling}\label{subsec:network modeling}
In MAM, local quality response is inherently neighborhood-dependent. Each fusing location is influenced by not only its own process observations, but also nearby scan tracks and adjacent layers with potentially heterogeneous geometric and process settings. These interactions are difficult to be fully captured when fusing locations are treated as independent samples or simple sequential observations. Hence, we propose to represent the MAM process as a complex network. 
 
\begin{figure*}[ht]
    \centering
    \includegraphics[width=\linewidth]{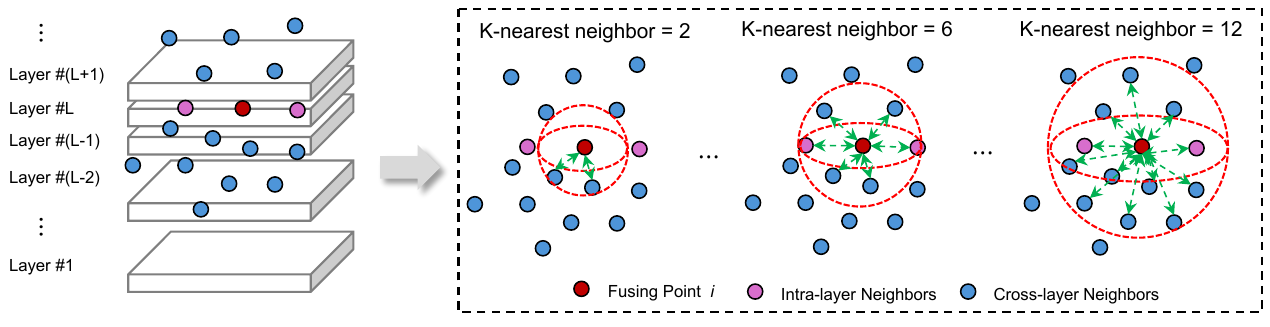}
    \caption{Illustration of k-nearest-neighbor-based graph construction.}
    \label{fig:knn}
\end{figure*}

In this network, each node corresponds to a local fusing point and each edge between two nodes encodes their physically meaningful relationships. One natural way to connect nodes is through a 3D ball structure, i.e., two nodes are connected if their spatial distance fell within a fixed radius (e.g., a multiple of the hatching space). While this strategy provides an intuitive way to capture process interactions, we observed that the number of adjacent nodes can increase substantially when the neighborhood radius expands from one hatching space to two hatching spaces. For example, the average node degree increases from 20.39 to 139.06 in the benchmark dataset utilized in this study. Such rapid growth in neighborhood size makes learning with graph transformers computationally prohibitive, particularly in our setting where both within-node attention and attention across neighboring nodes must be considered. We instead employ a $k$-nearest-neighbor-based graph construction strategy (see Figure \ref{fig:knn}), which maintains a sparse and computationally tractable network while preserving meaningful spatiotemporal relationships among process observations. 

Specifically, if two nodes do not belong to each other's $k$ nearest neighbor sets, no edge is established between them and their corresponding edge weight is set to 0. For each valid node pair $(i,j)$ under the $k$-nearest-neighbor-based graph construction, the edge weight $w_{ij}$ is modeled as the product of a spatial proximity term $\kappa_s(i,j)$
\begin{equation}
\kappa_s(i,j) = \exp\!\left(-\frac{\left\lVert \mathbf{s}_i-\mathbf{s}_j \right\rVert_2}{a\,h^{2}}\right) 
\end{equation}
and a process dependency term $\kappa_p(i,j)$
\begin{equation}
\kappa_p(i,j) = \exp\!\left(-\frac{1}{2}(\mathbf{p}_i-\mathbf{p}_j)^{\top}\mathbf{\Sigma}^{-1}(\mathbf{p}_i-\mathbf{p}_j)\right)
\end{equation}
where the $l_2$-norm $\left\lVert \mathbf{s}_i-\mathbf{s}_j \right\rVert_2$ denotes the 3D spatial distance between node $i$ and $j$ located at $\mathbf{s}_i$ and $\mathbf{s}_j$, respectively. The constant $a$ is introduced to control the influence of hatching space $h$ on the spatial proximity term. In addition, $\kappa_p(i,j)$ is a multivariate Gaussian kernel to capture dependencies across process parameters (e.g., laser power, scanning velocity). Here, $\mathbf{p}_i$ denotes the process parameter vector associated with node $i$ and $\mathbf{\Sigma}$ denotes the covariance matrix of the multivariate process parameters.

Collecting all pairwise weights yields a sparse weighted adjacency matrix $W=[w_{ij}]\in\mathbb{R}^{N\times N}$ for the spatiotemporal network, where $w_{ij} = \kappa_s(i,j) \times \kappa_p(i,j)$ for valid node pairs and $N$ denotes the total number of fusing points. This matrix provides a quantitative representation of process coupling throughout the 3D build. By restricting nonzero entries to physically meaningful local neighborhoods and modulating their magnitudes through spatial and process similarity, $W$ preserves the underlying build context and establishes a principled foundation for downstream learning on the spatiotemporal process network. 
\subsection{Spatiotemporal Graph Transformer}\label{subsec:}
Based on the network structure, we further develop a dual-attention graph transformer to capture the effects of spatiotemporal neighborhood interactions on part quality. This architecture is designed for in-situ quality prediction and monitoring. When a new layer is completed, the framework predicts its quality state based on spatiotemporal neighborhood interactions both within- and cross-layers. 

\subsubsection{Input}
Instead of taking the entire spatiotemporal network as input, the proposed transformer operates on the 1-hop neighborhood graph of node $i$, denoted by $G_{i}^{(1)}$. Specifically, $G_{i}^{(1)}$ is a subgraph consisting of node $i$ and all nodes directly connected to it by one edge in the original network. Such a design is conceptually inspired by how Transformers are applied in natural language processing. In this analogy, each 1-hop neighborhood graph can be interpreted as a paragraph extracted from a larger document, while each node within the neighborhood corresponds to an individual sentence represented by its associated feature vector. Transformer then learns contextual dependencies among neighboring nodes in a manner analogous to how language models capture semantic relationships among sentences within a paragraph. By restricting attention to localized neighborhoods, the model can focus on the most relevant interaction while still leveraging the expressive power of attention mechanisms to encode spatiotemporal interactions. 

\begin{figure*}[ht]
    \centering
    \includegraphics[width=\linewidth]{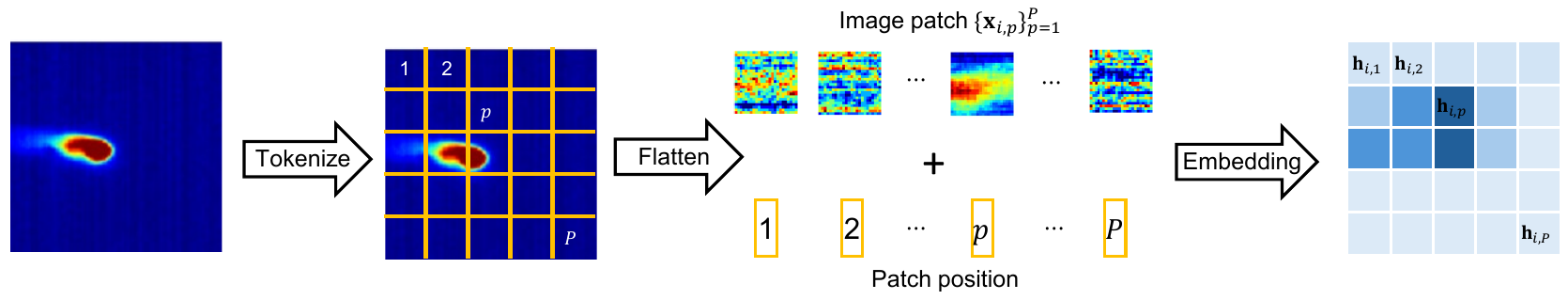}
    \caption{Patch-level image tokenization and embedding of node-level melt pool images.}
    \label{fig:image2token}
\end{figure*}

Additionally, each node $i$ in $G_{i}^{(1)}$ is characterized by sensing data collected at the corresponding fusing location. In this study, we focuses on melt-pool imaging $\mathcal{X}$ as the primary sensing modality. Accordingly, each node encodes localized quality-related features, whereas the 1-hop neighborhood graph captures how these features interact over the spatiotemporal neighborhood structure. As illustrated in Figure \ref{fig:image2token}, each node feature is first tokenized into a sequence of image patches, i.e., $\{\textbf{x}_{i,p}\}_{p=1}^{P}$, where $P$ is the total number of patches. In this way, each node is represented as an ordered collection of patch-level tokens that preserve local sensing details while allowing the model to learn both localized image characteristics within an individual node and interactions among neighboring nodes in the graph. 

Unlike a conventional transformer, our model incorporates positional encoding (PE) at two hierarchical levels to reflect the nested structure of the input. The first level is patch PE, which preserves the spatial location of each patch within a node image. It is determined by patch location $p$ using sinusoidal functions with different frequencies \citep{vaswani2017attention},  
\begin{equation}
    \mathrm{PE}(p, 2k) = \sin\left(\frac{p}{10000^{2k/d}}\right) 
\end{equation}
\begin{equation}
    \mathrm{PE}(p, 2k + 1) = \cos\left(\frac{p}{10000^{2k/ d}}\right)
\end{equation}
where $k = {0, 1, ..., d/2-1}$ indexes the sinusoidal frequency pairs and $d$ is the token embedding dimension. Utilizing both sine and cosine functions give each frequency a 2D phase representation of the position. To maintain spatial consistency across nodes, patch PEs are shared so that patches at the same relative image position carry the same spatial meaning. Then, these PEs are added to the patch embeddings,
\begin{equation}
    \textbf{h}_{i,p} = f_\text{emb}(\textbf{x}_{i,p}) + \textbf{e}_{p}^{\text{patch}}
\end{equation}
where the function $f_\text{emb}$ maps the tokenized image patch into the embedding space and $\textbf{e}_{p}^{\text{patch}}$ denotes the embedded patch PE for dimensional consistency. Ultimately, each image is transformed into an ordered set of input patch tokens. 

The second level is node PE, which encodes the structural context of the node within the MAM process network. It is worth noting that the node's structural context is not merely a function of its spatial location in the 3D space, but rather its position within the underlying non-Euclidean network space. Mathematically, node PE can be characterized by the spectral graph analysis. Assume that $D$ is the degree matrix of the original process network and $W$ is its weighted adjacency matrix. Then, the Laplacian matrix is 
\begin{equation}
    L = D - W
\end{equation}
which encodes the network connectivity while preserving weighted relationships among nodes. Since the Laplacian $L$ is a real symmetric matrix, it can be eigendecomposed in the same way as a traditional symmetric matrix. 
\begin{equation}
    L \mathbf{u} = \lambda \mathbf{u}
\end{equation}
where $\mathbf{u}$ corresponds to the obtained eigenvectors. In general, eigendecomposition provides an orthogonal basis for representing the structure of a matrix. When applied to the Laplacian, the resulting eigenvectors can be viewed as smooth structural coordinates defined on the nodes of the network, with connected nodes tending to have similar values. By retaining the first several eigenvectors, each node is assigned a vector of spectral coordinates that reflect its position within the global network topology. These coordinates naturally serve as node PE, which carries topology-aware structural information for downstream representation learning. 

\subsubsection{Learning via Dual Attention}
\begin{figure}[ht]
    \centering
    \includegraphics[width=0.75\linewidth]{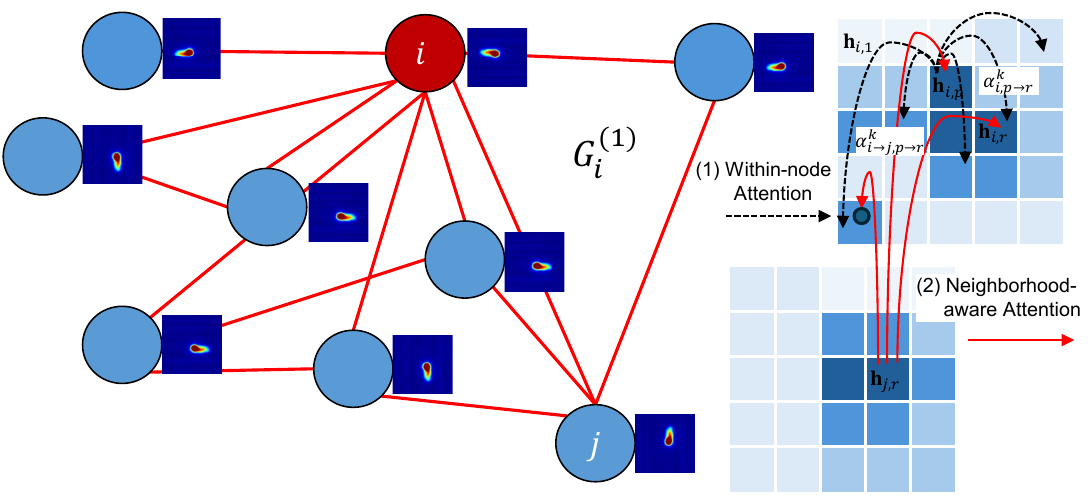}
    \caption{The dual-attention mechanism over the 1-hop neighborhood graph $G_i^{(1)}$ of node $i$: (a) within-node attention models interactions among image patches inside a node and (b) neighborhood-aware attention aggregates information from image patches of neighborhood nodes in the spatiotemporal graph. Note that only a subset of patch-wise connections is shown for visual clarity.}
    \label{fig:dual-attention}
\end{figure}

The 1-hop neighborhood input provides a structured representation of local spatiotemporal neighbors, including node-level patch tokens and PEs. Nonetheless, effective learning of process-quality relationships requires modeling image patch interactions not only within each node but also across neighboring nodes. To address this, we design a dual-attention mechanism (see Figure \ref{fig:dual-attention}) into the transformer block to model spatiotemporal neighborhood interactions and investigate how they support in-situ quality monitoring. Each transformer block has two attention levels: (1) within-node attention to extract local quality-related features and (2) neighborhood-aware attention to propagate and aggregate these local features over the neighborhood. 

Inside each transformer block, node-level patch embedding $\mathbf{h}_{i,p}$ is first updated via multi-head self-attention to model interactions with other patches within the same node, i.e.,
\begin{equation}
    \Tilde{\mathbf{h}}_{i,p} = W_{O, \text{self}} \left( \big\|_{k=1}^{H} \text{head}_{i,p}^{k} \right)
\end{equation}
where $W_{O, \text{self}} \in \mathbb{R}^{d \times Hd_h}$ denotes the output projection matrix that fuses information across all attention heads and transform the hidden representation into dimension $d$. $\big\|$ is the concatenation operator and $\text{head}_{i,p}^{k} \in \mathbb{R}^{d_h}$ represents the output of the $k$-th attention head for patch $p$. In fact, $\text{head}_{i,p}^{k}$ is a weighted aggregation of value vectors from all patches $r$ within the same node, 
\begin{equation}
    \text{head}_{i,p}^{k} = \sum_{r=1}^{P} \alpha_{i,p \to r}^{k} \mathbf{v}_{i, r}^{k}
\end{equation}
where $\mathbf{v}_{i, r}^{k} = W_{V, \text{self}}^{k} \mathbf{h}_{i,r}$ is the value vector of patch $r$ under head $k$ and $\alpha_{i,p \to r}^{k}$ denotes the attention weight from patch $p$ to patch $r$, 
\begin{equation}
    \alpha_{i,p \to r}^{k} = \frac{\text{exp} \left( \frac{(\mathbf{q}_{i, p}^{k})^{\top}\mathbf{k}_{i, r}^{k}}{\sqrt{d_{h}}}\right)}{\sum \limits_{r' = 1}^{P} \text{exp} \left( \frac{(\mathbf{q}_{i, p}^{k})^{\top}\mathbf{k}_{i, r'}^{k}}{\sqrt{d_{h}}}\right)}
\end{equation}
with $\mathbf{q}_{i, p}^{k} = W_{Q, \text{self}}^{k} \mathbf{h}_{i,p}$ and $\mathbf{k}_{i, r}^{k} = W_{K, \text{self}}^{k} \mathbf{h}_{i,r}$ correspond to the query and key representations, respectively. Similar to $W_{V, \text{self}}^{k}$, both $W_{Q, \text{self}}^{k}$ and $W_{K, \text{self}}^{k}$ are learnable projection matrices. The scaled dot product of the query and key representations $\frac{(\mathbf{q}_{i, p}^{k})^{\top}\mathbf{k}_{i, r}^{k}}{\sqrt{d_{h}}}$ measures the relevance of patch $r$ to patch $p$. In our context, it reflects how strongly local characteristics captured by patch $r$ contribute to refining the representation of patch $p$, which is conducive to identifying informative within-node patterns associated with process and quality variations. 

After within-node attention, a feed-forward subnetwork is applied to refine individual patch embeddings through nonlinear transformation. This subnetwork is consisted of a pre-normalized layer and two linear layers separated by a GELU nonlinearity. Instead of directly operating on the after-attention result, the pre-normalized layer performs normalization on the addition of after-attention and before-attention results via a residual connection (i.e., $\Tilde{\mathbf{h}}_{i,p} + \mathbf{h}_{i,p}$). The normalized result is then projected into a larger hidden space via the first linear layer. Compared with ReLU, GELU provides a smoother nonlinear transformation by introducing the standard Gaussian CDF, 
\begin{equation}
    \text{GELU}(x) = x \Phi(x)
\end{equation}
and is therefore preferred in Transformer architectures. After GELU, the final layer performs linear transformation again so as to restore the original hidden dimension. 

Based on the updated patch-level representations within each node, neighborhood-aware attention seeks to learn how these representations should be propagated across the spatiotemporal neighborhood $\mathcal{N}(i)$. Similar to within-node attention, neighborhood-aware attention leverages key and query representations of image patches to estimate attention scores between pairs of patches. However, unlike within-node attention, the neighborhood-aware attention score quantifies the relevance between patch $p$ in node $i$ and patch $r$ in a neighboring node $j \in \mathcal{N}(i)$, 
\begin{equation} \label{eq: cross-attention score}
    \alpha_{i \to j,p \to r}^{k} = \frac{\text{exp} \left( \frac{(\mathbf{q}_{i, p}^{k})^{\top}\mathbf{k}_{j, r}^{k}}{\sqrt{d_{h}}}\right)}{\sum \limits_{j' \in \mathcal{N}(i)} \sum \limits_{r' = 1}^{P} \text{exp} \left( \frac{(\mathbf{q}_{i, p}^{k})^{\top}\mathbf{k}_{j', r'}^{k}}{\sqrt{d_{h}}}\right)}
\end{equation}
It is important to note that the normalization is performed over all $(\lvert \mathcal{N}(i) \rvert \, P)$ candidate neighboring patches, where $\lvert \mathcal{N}(i) \rvert$ denotes the number of neighbors associated with node $i$ and $P$ is the total number of patches within one node. This quantity depends on both the constructed network structure and the intended model usage. For example, although cross-layer neighbors may in principle come from both previous and subsequent layers, node features (i.e. in-situ sensing data) from subsequent layers are not yet available during online monitoring. As a result, future nodes are excluded from the neighborhood definition for in-situ monitoring tasks. 

In addition, to account for node positional information, node PE derived from the graph spectral analysis is incorporated into its patch-level representation prior to neighborhood-aware attention. For node $i$, the input to neighborhood-aware attention is defined as
\begin{equation}
    \Tilde{\Tilde{\mathbf{h}}}_{i,p} = g_\text{emb} (\mathbf{u}_i) +  \Tilde{\mathbf{h}}_{i,p}
\end{equation}
where $g_\text{emb}$ maps the node PE $\mathbf{u}_i$ into a consistent dimension space and $g_\text{emb}$  $\Tilde{\mathbf{h}}_{i,p}$ denotes the patch-level representation after self-attention and subsequent feed-forward subnetwork. As a result, $\mathbf{q}_{i, p}^{k}$ and $\mathbf{k}_{j,r}^{k}$ in Equation \ref{eq: cross-attention score} are the linear projections of $\Tilde{\Tilde{\mathbf{h}}}_{i,p}$ in the query and key space, respectively. The cross-attention output of head $k$ is therefore the aggregation of all candidate patches, 
\begin{equation}
    \text{head}_{i,p, \text{cross}}^{k} = \sum \limits_{j \in \mathcal{N}(i)} \sum \limits_{r = 1}^{P} \alpha_{i \to j,p \to r}^{k} \mathbf{v}_{j, r}^{k}
\end{equation}
Through joint learning across $H$ attention heads, patch-level representations are further refined to incorporate the aggregated information from neighboring nodes 
\begin{equation}
    \hat{\mathbf{h}}_{i,p} = W_{O, \text{cross}} \left( \big\|_{k=1}^{H} \text{head}_{i,p, \text{cross}}^{k} \right)
\end{equation}
In this way, each patch representation captures not only within-image dependencies but also inter-node interactions. Finally, $\hat{\mathbf{h}}_{i,p}^{l}$ is passed through another feed-forward subnetwork for nonlinear refinement. 

By stacking multiple transformer blocks, the model progressively learns richer patch-level representations that encode spatiotemporal neighborhood interactions. The resulting features are then pooled and passed through a projection head to estimate the target quality variable, i.e., 
\begin{equation}
        \hat{\textbf{y}}_i = f_{\text{pred}} \left( \text{Pool} \left(\{\hat{\mathbf{h}}_{i,p}^{l}\}_{p=1}^{P} \right) \right)
\end{equation}
where $\text{Pool}(\cdot)$ denotes a pooling operation over patch-level representations, and $f_{\text{pred}}(\cdot)$ denotes a projection head such as a linear layer and multilinear perceptron. Depending on the formulation and the design of the projection head, the output may be either a scalar or a vector. 

\section{Experimental Design}{\label{sec:exp design}}
\subsection{Dataset}
\begin{figure}[h]
    \centering
    \includegraphics[width=0.5\linewidth]{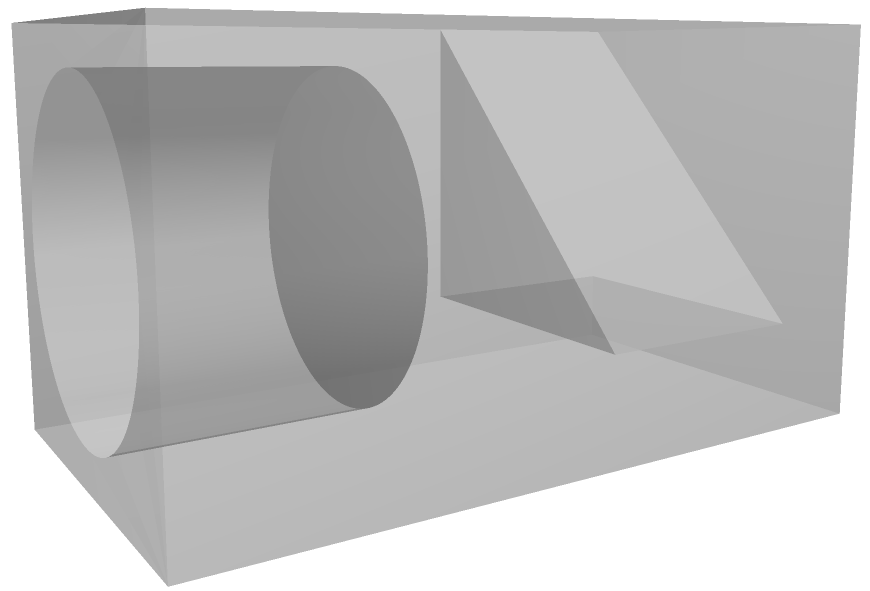}
    \caption{3D geometry of the overhang part}
    \label{fig:overhang part}
\end{figure}

The proposed framework was evaluated on the NIST AMS 100-69 benchmark dataset \citep{lane2020process, yang2025fully}, which is a process monitoring dataset with metal powder bed fusion additive manufacturing for fabricating overhang parts. The part has an overhang geometry with a horizontal cylindrical cutout and overall dimensions are 9 mm $\times$ 9 mm $\times$ 5 mm (see Figure \ref{fig:overhang part}). It consists of 250 layers fabricated using a serpentine scan pattern. The controlled process parameters includes laser scan speed, power and locations. In particular, the laser is scanned at a speed of 900 mm/s for the contour region and 800 mm/s for the infill region, whereas the laser powers are set to be 100 W and 195 W for the contour and infill regions, respectively. 

This study focuses on three data files: (1) digital commands from the scan file that specify laser positions and controlled settings, (2) melt pool monitoring images captured by a co-axial high-speed camera, which are digital records of localized process states during fabrication, and (3) Micro-XCT scan data, which provides ex-situ measurements of the quality conditions of the fabricated parts. The objective of this study is to evaluate the proposed framework's capability to integrate multimodal information for quality prediction and in-situ quality analysis. 
 
The experiments were conducted on 16 consecutive layers, which comprises a total of 88,825 fusing locations. Each fusing location is characterized by its design, process, and quality information. The design information includes the 3D spatial location and commanded laser power, which are utilized to construct the manufacturing process network. The process information primarily consists of melt-pool imaging data, which serves as node features in the constructed network. The quality information corresponds to registered ex-situ quality measurements, which serves as the model outputs/targets. The dataset was partitioned into training, validation and testing subsets according to layer indices. Specifically, layer 31-38 were utilized for model training, layer 39-42 for validation, and layer 43-46 for testing. 

\subsection{Experimental Details}
\begin{figure}[ht]
    \centering
    \includegraphics[width=0.75\linewidth]{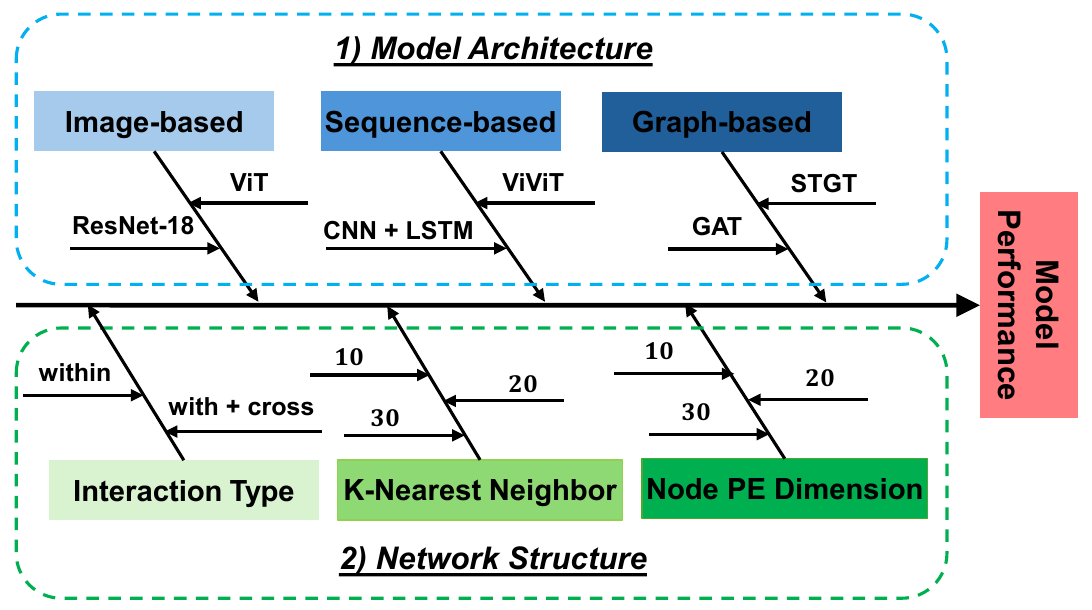}
    \caption{Diagram of the experimental design.}
    \label{fig:fishbone diagram}
\end{figure}

As shown in Figure \ref{fig:fishbone diagram}, the experimental design is structured around two primary components: model architecture and network structure. The first component focuses on comparing the proposed framework against a range of baseline models including image-based, sequence-based, and graph-based approaches. Each category includes one CNN-based model and one Transformer-based model. This comparison is designed to evaluate whether the proposed framework can more effectively capture process-quality relationships by modeling spatiotemporal neighborhood interactions. The second component investigates the effect of network structure design, such as interaction type, neighborhood construction, and node positional encoding on the predictive capability of the proposed framework. 

For model training, we employ the Adam optimizer with different decoupled weight decay values and learning rates based on model categories. For example, CNN-based models adopt a decoupled weight decay of 0.0001 and a learning rate of 0.001, whereas the remaining considers a decoupled weight of 0.001 and a learning rate of 0.0005. The detailed model architectures are summarized as follows, 
\begin{itemize}
    \item \textit{CNN-based}: The image-based model is built upon ResNet-18, which introduces skip connections into a deep convolutional neural network of 18 layers \citep{he2016deep}. The sequence-based model employs a CNN encoder with an output dimension of 256, followed by a bidirectional LSTM with 2 layers and 128 hidden units in each direction. The graph-based model is based on a graph attention network (GAT) \citep{velivckovic2017graph}. 
    \item \textit{Transformer-based}: The image-based model is based on the Vision Transformer (ViT) \citep{dosovitskiy2020image}, whereas the sequence-based model is built on the Video Vision Transformer (ViViT) \citep{arnab2021vivit}. For all Transformer-based models, the patch size is fixed at 24, the number of attention heads is set to 4, the feed-forward subnetwork's hidden dimension is set to 128, and the model depth is set to 4. 
\end{itemize}
Finally, model performance is evaluated using two statistical measures, i.e., root mean square errors (RMSE) to quantify the overall model predictivity and $R^2$ to quantify the proportion of explained variation by different analytical models. 

\section{Experimental Results} \label{sec:results}
We first compared the performance of different CNN-based and Transformer-base models for predicting the quality variable under the assumption that only within-layer interactions are present. Table \ref{tab:overall_model_comparison} shows that predictive performance improves as the model becomes more faithful to represent the underlying data structure and incorporate stronger inductive bias. For example,  the $R^2$ value increases from 0.386 to 0.440 to 0.451 when moving from ViT to ViViT to STGT. This shows that ViViT benefits from learning process-quality relationships from scanning sequences rather than treating observations independently as in ViT, and STGT further enhances the performance by introducing process-informed network structure and explicitly modeling spatiotemporal neighborhood interactions. At the same time, CNN-based models show a similar improvement pattern from image-based to sequence-based to graph-based models. When putting CNN-based and transformer-based models together, their predictive performances are comparable to each other under similar data representations. 
\begin{table}[ht]
\centering
\caption{Performance comparison of model architectures on the test dataset assuming that only within-layer interactions are allowed and the sequence of neighboring history is 20 if applicable.}
\label{tab:overall_model_comparison}
\begin{tabular}{llcc}
\hline
Model Type & Architecture & RMSE & \(R^2\) \\
\hline
\multirow{3}{*}{CNN-based} & ResNet-18 & 0.0228 & 0.389 \\
& CNN + LSTM & 0.0216 & 0.451 \\
& GAT & 0.0211 & 0.471 \\ \hline 
\multirow{3}{*}{Transformer-based} & ViT & 0.0229 & 0.386 \\
& ViViT & 0.0218 & 0.440 \\
& STGT (ours) & 0.0215 & 0.451 \\
\hline
\end{tabular}
\end{table}

\begin{figure}[ht]
    \centering
    \includegraphics[width=0.6\linewidth]{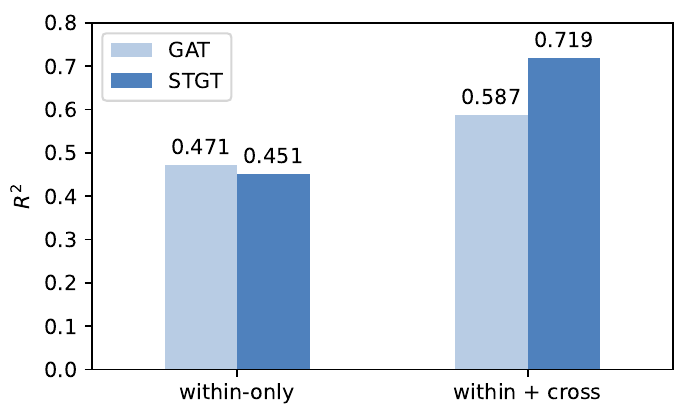}
    \caption{Comparison of $R^2$ values for GAT and STGT under two interaction settings.}
    \label{fig:interaction type}
\end{figure}

We then conducted experiments to test the hypothesis that incorporating cross-layer interaction is significant for capturing process-quality relationships. Specifically, the first experimental setting considers only within-layer interactions, whereas the second allows both within-and cross-layer interactions. Although the number of neighbors is controlled to be the same in both settings, the difference in the interaction type leads to different network structures including different edges, edge weights, and potentially node positional encodings. Figure \ref{fig:interaction type} compares the performance of GAT and STGT under these two settings. The results show that the second setting consistently yields better performance regardless of model architecture. For GAT, the $R^2$ value increases from 0.471 to 0.587. For STGT, it increases more substantially from 0.451 to 0.719. This can be largely attributed to the introduction of node positional encoding as additional structural details. Overall, these results show that cross-layer interactions provide important contextual information for learning process-quality relationships and suggest that STGT is more effective than GAT in leveraging such spatiotemporal dependencies. 

\begin{figure}[ht]
    \centering
    \includegraphics[width=\linewidth]{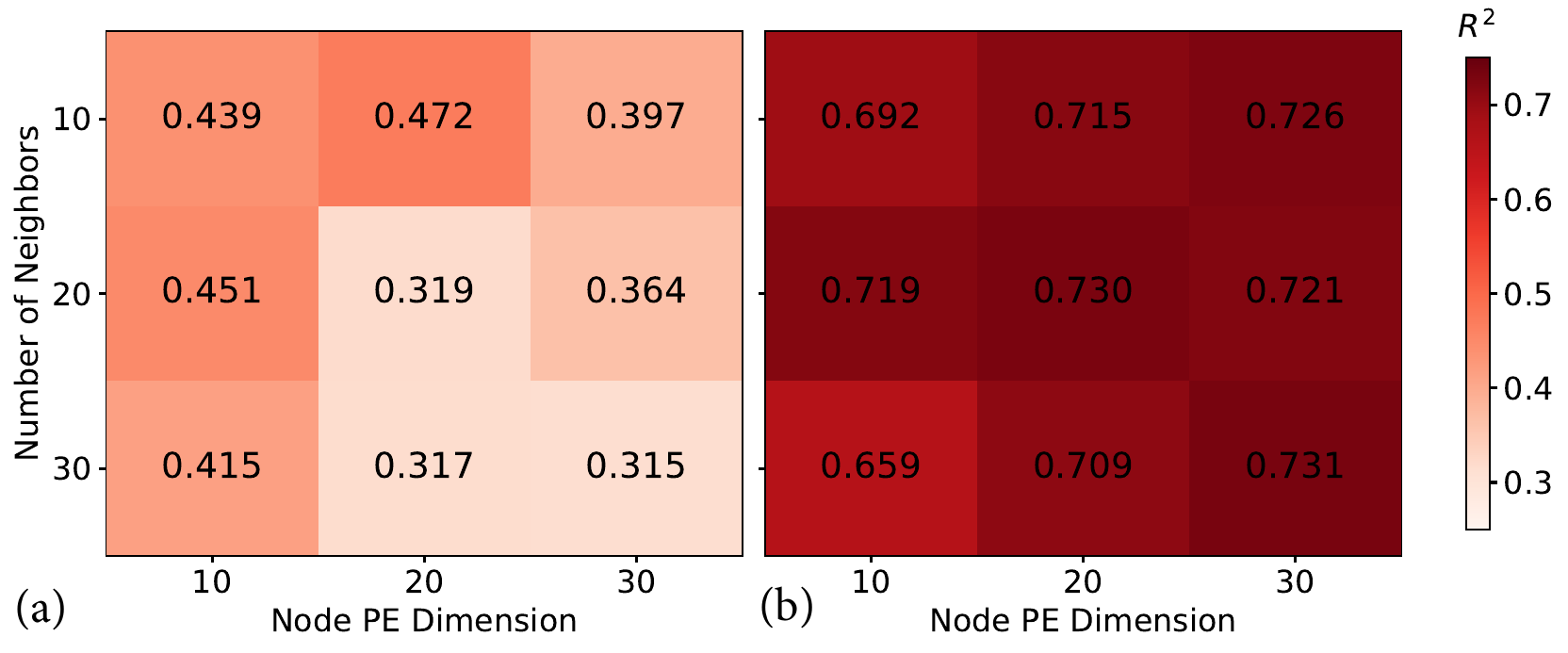}
    \caption{Sensitivity of STGT to neighborhood size and node PE dimension under (a) the within-layer setting and (b) the within-and cross-layer setting.}
    \label{fig:paramter sensitivity}
\end{figure}

In addition, we investigated the influence of neighborhood size and node PE dimension on the performance of STGT. The heatmap in Figure \ref{fig:paramter sensitivity}(a) show the results under the within-layer setting, where only intra-layer interactions are considered. Under this setting, the $R^2$ metric is quite sensitive to the number of neighbors utilized in graph construction and generally decreases as more neighbors are included. This shows that enlarging the neighborhood may introduce redundant or less informative connections that dilute useful local information. Among the tested node PE dimensions, a value of 10 tends to yield better performance than the other settings. 

In contrast, Figure \ref{fig:paramter sensitivity}(b) presents the results under the within-and cross-layer setting, where both intra-and inter-layer interactions are incorporated. In this setting, a larger node PE dimension of 30 tends to achieve better prediction accuracy. This shows that richer positional encoding is beneficial when the graph structure becomes more complex due to the introduction of cross-layer dependencies. Moreover, the influence of neighborhood size on model performance becomes less pronounced than in the within-layer case. Overall, these results show that the optimal network design depends strongly on the interaction setting. 

Moreover, we analyzed the learned neighborhood attention weights to gain physical insights into the modeled three-dimensional process interactions. Each neighborhood attention weight describes the relative importance assigned by a query node to one of its neighboring nodes averaged across attention heads and query patches. Since a neighbor may come from either the same layer or previously deposited layer, we define a layer offset variable to describe the layerwise distance between the query node and its neighbor. A layer offset of 0 indicates that the neighbor is located in the same layer, whereas a layer offset of 4 indicates that the neighbor is four layers away from the query node. 

\begin{figure}[ht]
    \centering
    \includegraphics[width=0.6\linewidth]{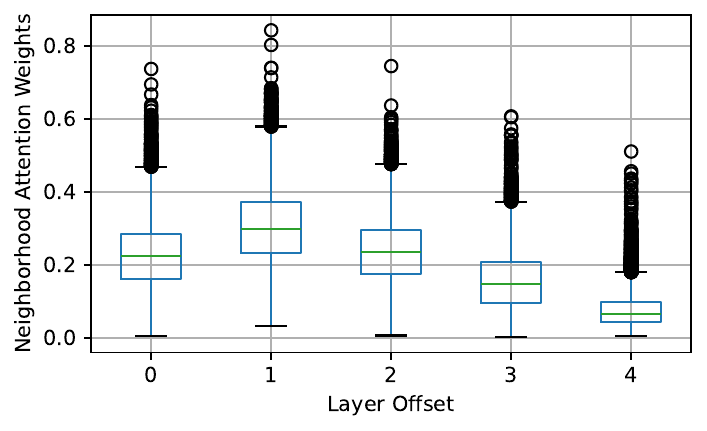}
    \caption{Distribution of learned neighborhood attention weights across layer offsets. The results are obtained from the STGT model with neighborhood size = 20 and node PE dimension = 20.}
    \label{fig:distribution of learned neighborhood attention weights}
\end{figure}

Figure \ref{fig:distribution of learned neighborhood attention weights} shows the distribution of neighborhood attention weights across different layer offsets. The green line inside each box represents the median and the whiskers refers to the range of non-outlier values. A clear pattern is that the median of the attention weight increases from layer offset 0 to layer offset 1, and then gradually decreases as the layer offset further increases. This suggests that neighbors from the immediately preceding layer contribute strongly to quality-relevant representation learning. The attention weights remain relatively high for neighbors with a layer offset of 2, suggesting that inter-layer effects may persist beyond the adjacent layer. As the layer offset becomes larger, the learned attention weights gradually decrease. This suggests a decay of cross-layer influence with increasing layerwise distance. The overall trend provides evidence that the proposed model does not treat all neighboring nodes equally. Instead, it learns a physically meaningful attention pattern in which recent cross-layer neighborhood history receives greater emphasis than more distant layers. 

\begin{figure*}[ht]
    \centering
    \includegraphics[width=\linewidth]{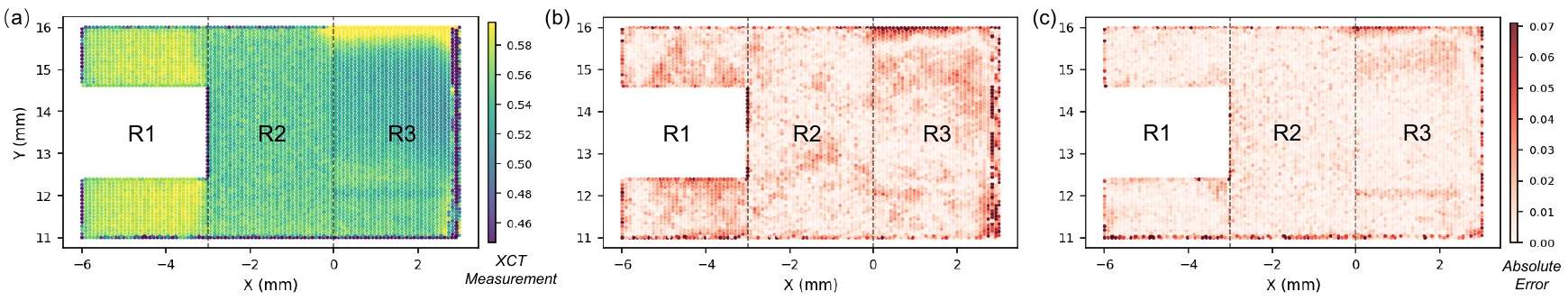}
    \caption{Two-dimensional spatial maps of (a) measured quality values and absolute prediction error of the STGT via (b) within layer and (c) within- and cross-layer neighborhood context for one test layer.}
    \label{fig: 2d spatial maps}
\end{figure*}

Finally, we analyzed model performance across different spatial regions of the part. As shown in Figure \ref{fig: 2d spatial maps}(a), the selected layer is divided into three regions. R1 represents a geometric-transition region, where layerwise geometry is captured within the selected layer range. R2 corresponds to a geometric-stable region, while R3 is associated with a geometric-transition region in the full part but has not yet undergone geometric variation in the current selected layer. The quality measurements from XCT are spatially heterogeneous across the two-dimensional layer due to geometric effects. Figure \ref{fig: 2d spatial maps}(b) and (c) compares the pointwise absolute prediction errors of STGT under the within-layer-only, and within-and-cross-layer settings. Compared with the within-layer-only setting, the second setting generally produce lower and more spatially uniform errors over most regions of the layer. This reduction is particularly visible in regions where within-layer model exhibits larger localized errors, which shows that incorporating cross-layer interactions helps the model better characterize process history and its influence on build quality. 

\begin{table}[ht]
\centering
\caption{Prediction accuracy of the within-layer and within-and cross-layer STGT across three geometric regions of the part.}
\label{tab:regional_rmse_comparison}
\footnotesize
\setlength{\tabcolsep}{3pt}
\begin{tabular}{c|c|c|c|c}
\hline
\multirow{2}{*}{Region} & \# Fusing & RMSE & RMSE & Improvement \\
& Locations & (within) & (within+cross) &  (\%) \\
\hline
R1 & 4,795  & 0.029 & 0.018 & 45.98 \\
R2 & 8,309  & 0.020 & 0.012 & 33.46  \\
R3 & 8,430  & 0.025 & 0.015 & 34.70 \\
\hline
\end{tabular}
\end{table}
Table \ref{tab:regional_rmse_comparison} further quantifies the region-wise improvement observed in Figure \ref{fig: 2d spatial maps}. Across all three regions, STGT with both within-and cross-layer interactions reduces RMSE compared with STGT using within-layer interactions only. The improvement is most pronounced in R1, where RMSE decreases from 0.029 to 0.018, corresponding to a 45.98\% reduction. R2 and R3 also show substantial reduction of 33.46\% and 34.70\%. Since R1 represents a geometric-transition region within the selected layer range, this result shows that cross-layer information is particularly beneficial in regions where local geometry changes across layers. Overall, experimental results demonstrate that explicitly incorporating cross-layer neighborhood interactions improves both global prediction accuracy and spatially localized prediction performance. This suggests the effectiveness of the proposed STGT framework for learning quality-relevant representations in MAM. 

\section{Conclusion} \label{sec:conclusions}
In layerwise fabrication processes like MAM, the quality at a given fusing location can be influenced by nearby locations within the same layer as well as previously deposited layers. These effects can be further complicated by heterogeneous part geometries, scan strategies, and process settings associated with different digital designs. However, many existing methods mainly model sensor data as independent samples or scan-path-based temporal sequences and tend to overlook cross-layer neighborhood interactions for quality analysis. 

This paper fills this gap by proposing a network-based representation of the MAM process, where fusing locations across the 3D build are connected through weighted relationships that reflect spatial proximity and process similarity. This representation also provides a structured way to integrate geometric information, process settings, and in-situ sensing data for quality prediction. Building on this representation, a spatiotemporal graph transformer is developed to learn quality-relevant features through within-node attention and neighborhood-aware attention. Experimental results from a real-world case study demonstrate incorporating 3D neighborhood interactions improves prediction performance compared with other state-of-the-art models. Additional analyses further shows that the learned attention patterns and region-wise error reductions provide useful insights into multi-layer process interactions. Future work will focus on extending the framework to different part geometries and developing more transferable frameworks for real-time monitoring and control systems. 

\backmatter
\bmhead{Acknowledgments}
The authors would like to thank the financial support from the Department of Industrial Engineering at Southern Illinois University Edwardsville.\\

\section*{Declarations}

\textbf{Conflict of interest} The authors declare no competing interests. \\

\noindent \textbf{Data availability} The data used in this study is publicly available at \url{http://doi.org/10.18434/M32233}.\\

\noindent \textbf{Code availability} Code will be made available on request. \\

\noindent \textbf{Author contribution} Joyce Karen Pelaez: data curation, methodology, validation, writing - original draft; Siqi Zhang: conceptualization, methodology, supervision, writing - original draft, review and editing; Hoo Sang Ko: supervision, writing - review and editing.

\noindent
If any of the sections are not relevant to your manuscript, please include the heading and write `Not applicable' for that section. 

\bibliography{AM}

\end{document}